\documentclass[lettersize,journal]{IEEEtran}
\usepackage{amsmath,amssymb}
\usepackage{algorithmic}
\usepackage{algorithm}
\usepackage{array}
\usepackage[caption=false,font=normalsize,labelfont=sf,textfont=sf]{subfig}
\usepackage{textcomp}
\usepackage{stfloats}
\usepackage{multirow}
\usepackage{url}
\usepackage{verbatim}
\usepackage{graphicx}
\usepackage{cite}
\usepackage{booktabs}
\usepackage{colortbl}
\usepackage{xcolor}
\usepackage[utf8]{inputenc}   
\usepackage[T1]{fontenc}      
\usepackage[normalem]{ulem}
\usepackage{lipsum}
\usepackage{array}  
\usepackage{xspace}
\usepackage{tikz}
\usepackage{hyperref}
\usepackage{xcolor} 
\begin{document}

\title{HSLiNets: Evaluating Band Ordering Strategies in Hyperspectral and LiDAR Fusion}

\author{{Judy X~Yang,~\IEEEmembership{Graduate Student Member,~IEEE},
        Jing~Wang,
        Zhuanfeng, Li,
        Chenhong Sui
       Zekun~Long,~\IEEEmembership{Graduate Student Member,~IEEE}, and Jun~Zhou,~\IEEEmembership{Senior Member,~IEEE} }
   } 



\maketitle

\begin{abstract}
The integration of hyperspectral imaging (HSI) and Light Detection and Ranging (LiDAR) data provides complementary spectral and spatial information for remote sensing applications. While previous studies have explored the role of band selection and grouping in HSI classification, little attention has been given to how the spectral sequence—or band order—affects classification outcomes when fused with LiDAR. In this work, we systematically investigate the influence of band order on HSI-LiDAR fusion performance. {Through extensive experiments, we demonstrate that band order significantly impacts classification accuracy, revealing a previously overlooked factor in fusion-based models. Motivated by this observation, we propose a novel fusion architecture that not only integrates HSI and LiDAR data but also learns from multiple band order configurations.} {The proposed method enhances feature representation by adaptively fusing different spectral sequences, leading to improved classification accuracy. Experimental results on the Houston 2013 and Trento datasets show that our approach outperforms state-of-the-art fusion models. 
Data and code are available at https://github.com/Judyxyang/HSLiNets.
}
\end{abstract}


\section{Introduction}
HSI captures rich spectral details across hundreds of contiguous spectral bands, enabling precise material discrimination. In contrast, LiDAR provides high-resolution elevation and structural data that enhance spatial feature extraction. The fusion of HSI and LiDAR data has gained significant attention in remote sensing due to its ability to leverage complementary spectral and LiDAR information for improved classification performance~\cite{gao2020survey}\cite{wang2021spectral}. This synergy has enabled breakthroughs in diverse applications, including land cover classification, urban mapping, and vegetation analysis~\cite{karim2023current}. 
Significant progress has been made with frameworks such as that introduced by Xu et al.~\cite{xu2017multisource}, which employs dual CNNs to concurrently process spectral-spatial features from HSI and elevation information from LiDAR, setting new benchmarks in the domain.

{However, the rich information contained in hyperspectral and LiDAR data also causes challenges for the classification task. For example, paired HSI and Lidar data often have high dimensionality which reduces the classification accuracy. In addition, redundancy of spectral and spatial information also exists in the fused data~\cite{liangpei2022data}. These redundancy includes not only similarity within hyperspectral bands, but also correlation between hyperspectral and LiDAR data. To address the issue of high dimensionality of hyperspectral image classification, band selection methods have been widely adopted to retain informative spectral bands while eliminating redundant ones~\cite{fu2022review}. For example, techniques such as mutual information-based selection~\cite{jain2022unsupervised}, genetic algorithms~\cite{singh2022enhanced}, and graph-based spectral clustering~\cite{ma2024joint} optimize band subsets for hyperspectral classification tasks. Deep learning frameworks, including convolution autoencoders (CAEs) and attention-based mechanisms~\cite{wang2020attend}, further improve band selection by learning discriminative spectral subsets within deep neural networks~\cite{ zheng2022effective}.} {However, this line of research focus on selecting bands for hyperspectral classification task only, without considering the task in the context of fusion with LiDAR. To address this issue, Yang et al.~\cite{yang2024lidar} first  propose a LiDAR-guided cross-attention network to select optimal bands for fusion task specifically. The insight from this  work reveals the relationship, especially correlation, among HSI bands and LiDAR channels has a significant impact on the fusion and classification. }

{Besides band selection, another perspective of exploring the relationship of channels within the multi-modal data is changing orders of sequence. While it is important, all existing works~\cite{balestra2024lidar} in hyperspectral and LiDAR fusion adopt the same natural sequence order of the data, struggling to fully exploit the potential performance of classification . As demonstrated by Guo et al.~\cite{guo2021feature}, local band position does have big influence on learning optimal features for classification. Therefore, our work goes a step further in investigating the impact of local band position by systematically evaluating different band sequence orders for the classification. Changing band sequence order changes the local positional relationship, and has potential to significantly affect classification performance. Specifically, we investigated four different band sequence orders for HSI-LiDAR fusion and investigating the impact on classification performance. Moreover, we propose a two branch symmetric network (HSLiNet) to fuse hyperspectral and LiDAR data for effective classification. HSLiNet adopts early fusion of two orders of HSI bands and LiDAR in each branch and the features extracted from two orders are then fused together to take advantage of each order's strength at classifying its subset of classes. Extensive experiments reveal that certain band order could be beneficial for classifying a certain subset of classes.}

The main contributions of this letter are summarized as follows:
\begin{enumerate}
    \item We systematically investigated the impacts of band sequence order on hyperspectral and LiDAR fusion task. Moreover, we found that different band order have different advantages for classifying different subset of classes.
    \item We propose a novel network called HSLiNet for hyperspectral and LiDAR fusion by two levels of fusion: fusing the two data modality in early fusion with different band orders and fusing the features extracted from different orders for harnessing the advantage of both band orders.
    \item Our HSLiNet outperforms all compared hyperspectral and LiDAR fusion methods and achieved state-of-the-art performance on two hyperspectral LiDAR fusion benchmark datasets.
\end{enumerate}
 
{The rest of this letter is organised as follows. Section~\ref{method} 
introduces the proposed methods. Section II describes comprehensive experiments based on Houston 2013 and Trento. Section III draws a conclusion.}

\section{Methodology}
{This section presents our method for systematically analysing the influence of hyperspectral band order in HSI-LiDAR fusion. The proposed HSLiNet for fusing HSI and LiDAR with different orders is described including overall framework and detailed design.}

\begin{figure*}[!ht]
\centering
\includegraphics[scale=0.12]{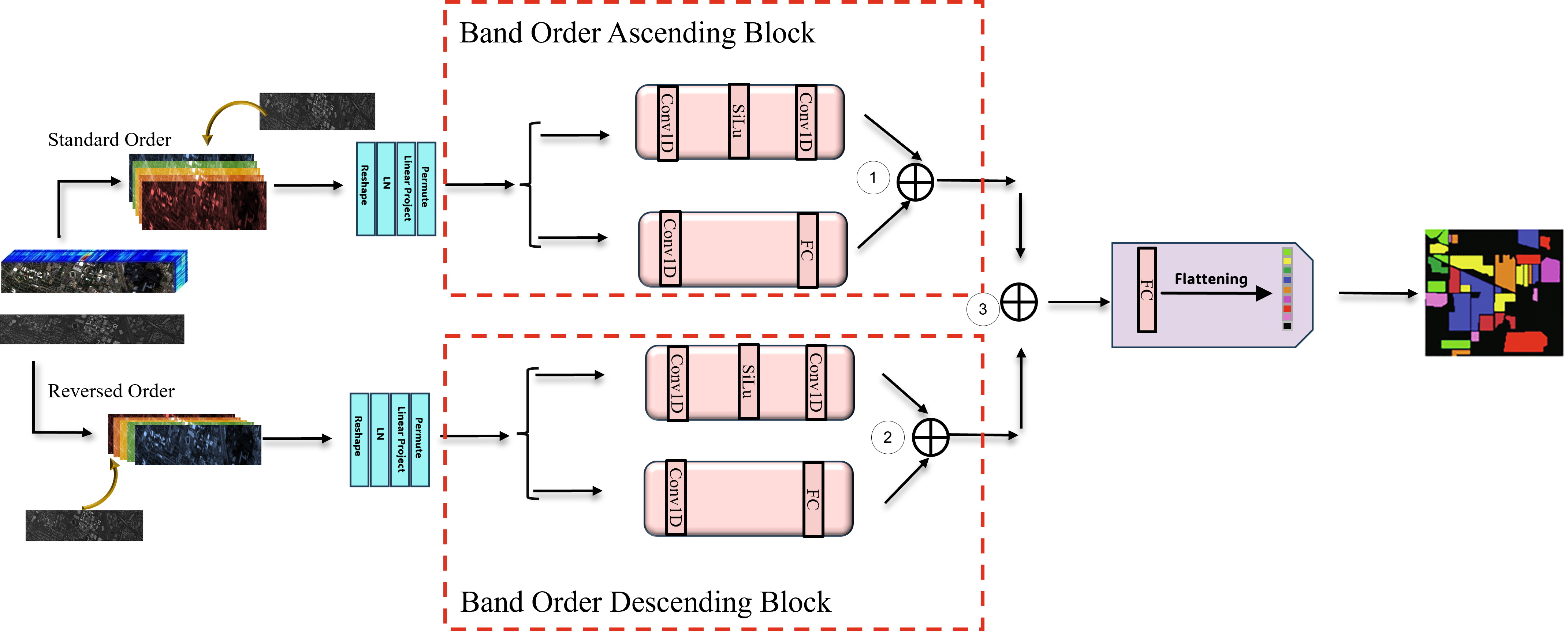}
\caption{Architectural overview of the proposed HSI-LiDAR fusion model. The framework consists of four main components: (A) Input patches( (HSI:  \( p \times p \times C \) and  LiDAR: \( p \times p \times 1 \) are processed through dual streams. (B)  B) Linear projection layers (\( \mathbb{R}^C \rightarrow \mathbb{R}^{d_h} \)) align modalities.(C) CNN blocks extract spectral-spatial features with band-order-specific parameters. (D) The fusion block combines streams before final classification.}
\label{hsilidvim}
\end{figure*}
\subsection{Proposed Method Overview and Architecture Description}
Fig.~\ref{hsilidvim} illustrates {a dual-stream hyperspectral and LiDAR fusion model designed for classification. The model incorporates two distinct band-order processing pipelines—ascending and descending—which are used to extract spectral features while integrating LiDAR data as an additional channel to provide complementary structural and elevation information. The extracted features from both band orders are fused and processed through a fully connected (FC) layer to generate the final classification output.}

{The architecture consists of three major components: input processing, band order feature extraction, and feature fusion with classification. The input processing stage takes in hyperspectral image (HSI) cubes and LiDAR elevation features. The hyperspectral data is denoted as $\mathbf{X_h} \in \mathbb{R}^{H \times W \times C}$, where {H} and {W} represent the spatial dimensions and {C} is the number of spectral bands.  Similarly, the LiDAR elevation map is represented as  $\mathbf{X_{li}} \in \mathbb{R}^{H \times W}$. To facilitate further processing, the spatial dimensions $H \times W$ 
are compressed into a single dimension $N$, where \( N = H \times W \) denotes the total number of pixels. Consequently, the hyperspectral data cube and LiDAR map are reshaped as follows:}
\[
\mathbf{X_h} \in \mathbb{R}^{N \times C}, \quad \mathbf{X_{li}} \in \mathbb{R}^{N \times 1}
\]

{This transformation simplifies subsequent analyses while ensuring that the spatial and spectral information remains aligned. Two band sequences of hyperspectral bands are used: the normal spectral band sequence and its reversed order.  These are processed independently, providing complementary spectral-spatial information. The dual-path design allows the model to learn diverse spectral patterns by processing both forward \( \mathbf{X_h}^{\rightarrow} \) (forward) and \( \mathbf{X_h}^{\leftarrow} \) (reversed) band sequences, ensuring robustness to variations in band order.}

{The individual band order feature extraction block is responsible for extracting spectral features from each band order configuration. The Band Order Ascending Fusion Block processes the spectral bands in their standard order, using 1D convolutional layers (Conv1D) followed by SiLU activation to extract meaningful spectral patterns. Fully connected (FC) layers further refine these feature representations. In contrast, the Band Order Descending Fusion Block follows a similar pipeline but processes the bands in reversed order. Both blocks function independently, ensuring that the model can capture spectral relationships regardless of the original sequencing of bands.}

{In the feature fusion and classification stage, the extracted features from both fusion blocks (Output 1 and Output 2) are combined via element-wise summation. This fusion mechanism ensures that complementary spectral information from both processing streams is preserved, reinforcing the robustness of the learned feature representations. The fused features are then processed through an additional fully connected (FC) layer, which maps them to classification outputs. The final result is a segmented classification map, where different colors represent different land cover types.}  

{The fused representation is computed as Eq.~(\ref{ch5-4})}:
\begin{equation}
\label{ch5-4}
\mathbf{h}_{\text{fusion}} = \Phi(\mathbf{h_{1}}^{\rightarrow} \oplus \mathbf{h_{2}}^{\leftarrow})
\end{equation}
where \( \oplus \) represents element-wise addition, and  \( \Phi \) denotes the FC layer, designed to refine the fused features. $\mathbf{h_{1}}^{\rightarrow}$ is the output from band order ascending block and $\mathbf{h_{2}}^{\leftarrow}$ is the output from band order descending block.

\subsection{Band Order Configurations}
To assess the impact of spectral band order on fusion performance, we introduce a dual-sequence strategy utilizing ascending and descending band orders. Specifically, we define four band ordering strategies: (1) Original order, where hyperspectral bands follow their natural spectral sequence; (2) Reversed order, where bands are processed in the opposite direction; (3) Descending importance order, ranking bands from most to least important based on the selection method in~\cite{yang2024lidar}; and (4) Ascending importance order, ranking bands from least to most important using the same method. LiDAR data is incorporated as an additional channel alongside each HSI sequence, enriching the feature space with structural and elevation information. These four configurations are analyzed to evaluate their influence on classification performance.

To fully utilize these spectral orderings, we employ a dual-pair fusion approach, where each pair consists of a primary and secondary spectral sequence. Pair 1 combines the original band order (primary) with its reversed version (secondary), while Pair 2 integrates the descending importance order (primary) with the ascending importance order (secondary). This approach systematically explores the model’s adaptability to spectral sequencing variations and its ability to extract complementary spectral-spatial patterns. By processing both forward and backward sequences in parallel, the model enhances robustness to spectral arrangement variations, improving classification accuracy.

\subsection{Classification and Loss Function}
{Once the fused feature representation is obtained, the final classification stage leverages a fully connected (FC) layer to generate class predictions. The model is optimised using the cross-entropy loss function, which is defined as Eq.~(\ref{ch5-5}):}

\begin{equation}
\label{ch5-5}
\mathcal{L} = \frac{1}{N}\sum_{i=1}^N \text{CE}(\mathbf{y}_i, \mathbf{\hat{y}}_i)
\end{equation}

{where, $N$: Total number of training samples, $\text{CE}$: Cross-entropy loss function, $\mathbf{y}_i \in \mathbb{R}^K$:  One-hot encoded ground truth label for $i$-th sample, $\mathbf{\hat{y}}_i \in \mathbb{R}^K$: Predicted class probabilities for $i$-th sample , and $K$ is Number of land-cover classes. }

\section{Experiments}
This section  describes experimental results of the impact of hyperspectral band order on classification performance and feature fusion.

\subsection{Experimental Setup}
Our framework is implemented by PyTorch, fusing raw HSI and LiDAR data before embedding. These fused patches are processed through non-linear directional blocks and spatial feature extraction modules. The experiments are conducted on an NVIDIA RTX 3090 GPU with 24GB VRAM. Training parameters setting include batch size of 32, learning rate of 0.00001, Adam optimizer, and 100 epochs. Table~\ref{tab:band_order_explanation} defines four hyperspectral band order configurations and their LiDAR variants.

\begin{table*}[ht!]
\centering
\caption{{Hyperspectral Band Order Configurations with LiDAR Integration}}
\label{tab:band_order_explanation}
\resizebox{\textwidth}{!}{
\begin{tabular}{|c|l|p{8cm}|}
\hline
\textbf{Data ID} & \textbf{Band Order Type} & \textbf{Description} \\
\hline
DB1 & Original Order & Hyperspectral bands arranged in their native wavelength sequence (e.g., 400–2500 nm). \\
\hline
DB2 & Reversed Order & Hyperspectral bands inverted from their original sequence (e.g., 2500–400 nm). \\
\hline
DB3 & Descending Importance Order & Selected bands ranked by importance (most → least discriminative). \\
\hline
DB4 & Ascending Importance Order & Selected bands ranked by importance (least → most discriminative). \\
\hline
DB1Li & Original Order + LiDAR & DB1 with LiDAR features (elevation, intensity) appended as pseudo-spectral bands at sequence start and end. \\
\hline
DB2Li & Reversed Order + LiDAR & DB2 with LiDAR features appended as pseudo-spectral bands. \\
\hline
DB3Li & Descending Importance + LiDAR & DB3 with LiDAR features appended as pseudo-spectral bands. \\
\hline
DB4Li & Ascending Importance + LiDAR & DB4 with LiDAR features appended as pseudo-spectral bands. \\
\hline
\end{tabular}}
\end{table*}

\subsection{Impact of Band Order on Classification}
 We analyse classification performance based on different hyperspectral band orders of two data sets under different settings. 
 
Table~\ref{pixel_summary} presents  classification performance using HSI single-pixel across different band orders. The experimental result shows that the reversed order (DB2) performed slightly better than the original band order (DB1). Specifically, OA values were 0.9191 for DB2 vs. 0.9110 for DB1. The impact was more significant for selected band orders (DB3, DB4), where the performance varied by up to 6\% . When integrating LiDAR, the result has been improved stability across all band orders. With LiDAR, Less-to-important band order (DB4Li) achieves the highest accuracy of 0.9277, demonstrating the robustness of fused spectral-spatial features. 

\begin{table*}[ht!]
\centering
\caption{{Pixel-Level Classification Performance Summary,Bold Data Represents the Best Method Experimental Results}}
\label{pixel_summary}
\resizebox{\textwidth}{!}{
\begin{tabular}{|l|c|c||c|c||c|c|c||c|c|c|}
\hline
Category & \multicolumn{4}{c||}{HSI Only} & \multicolumn{6}{c|}{HSI + LiDAR} \\
\hline
 & DB1 & DB2 & DB3 & DB4 & DB1Li & DB2Li & DB1Li+DB2Li & DB3Li & DB4Li & DB3Li+DB4Li \\
\hline
OA & 0.9110 & 0.9191 & 0.9086 & 0.9155 & 0.9459 & 0.9469 & \textbf{0.9568} & 0.9443 & 0.9452 & 0.9482 \\
AA & 0.9167 & 0.9237 & 0.9140 & 0.9193 & 0.9481 & 0.9467 & \textbf{0.9575} & 0.9440 & 0.9490 & 0.9484 \\
Kappa & 0.9035 & 0.9122 & 0.9008 & 0.9082 & 0.9412 & 0.9423 & \textbf{0.9531} & 0.9395 & 0.9405 & 0.9437 \\
\hline
\end{tabular}}
\end{table*}

Table~\ref{summary_patch9_s10} shows the band order impacting the classification performance using 10 training samples. The category-level and overall accuracy result can be seen variance based on different band orders. Experimental result illustrates DB1Li+DB2Li with OA of 0.8739 and DB3Li+DB4Li of 0.8558 outperform single-order fused configurations DB3Li with OA of 0.8388. Dual-path processing of complementary band orders maximises feature diversity, critical for fewer-data scenarios. 
For HSI-Only Baselines (DB1–DB4), Classes 8, 10, and  7 are low-performance classes. Class 8 (Lowest) is the worst in DB2 (0.4562), best in DB1Li+DB2Li (0.6345), improving 17.83\%. LiDAR fusion mitigates spectral ambiguity in highly challenging classes. In the fusion task, best in DB1Li+DB2Li (0.6204), improving 20.96\%, it indicates fusion leverages elevation data to resolve spectral confusion.  In Class 7,  DB1Li+DB2Li (0.9269) improves 28.22\% compared with DB4 (0.6447), because the fusion dramatically enhances performance for noise-prone classes with limited samples.
Notably, DB1Li+DB2Li (OA0.8739) and DB3Li+DB4Li (OA0.8558) outperform single-order fused configurations DB3Li (OA) 0.8388. 



\begin{table*}[ht!]
    \centering
     \caption{{Experiments Summarisation Comparison Based on Different Band Orders Using 10 Samples of Houston 2013 }}
 \resizebox{\textwidth}{!}{ 
    \begin{tabular}{|l|c|c|c|c||c|c|c|c|c|c|}
      \hline
        Class& \multicolumn{4}{c||}{Based on HSI} & \multicolumn{6}{c|}{Based on HSI + LiDAR} \\
        \cline{2-11}
        & DB1 &  DB2 & DB3 & DB4 & DB1Li &DB2Li& DB1Li+DB2Li & DB3Li & DB4Li &DB3Li+DB4Li \\
  \hline
        Class1  & \texttt{0.9758} & 0.9774 & 0.9613 & 0.9750 & 0.9629 & 0.9790 & 0.9637 & 0.9790 & 0.9484 & 0.9573 \\
        Class2  & 0.8264 & 0.8376 & 0.8368 & 0.8280 & 0.8296 & 0.8352 & 0.8481 & 0.8280 & 0.8328 & 0.8432 \\
        Class3  & 0.9956 & 0.9971 & 0.9971 & 0.9942 & 0.9971 & 0.9956 & 1.0000 & 0.9971 & 0.9971 & 0.9927 \\
        Class4  & \textbf{0.7626} & \textbf{0.8630} & \textbf{0.8695} & \textbf{0.8225} & 0.8995 & 0.9141 & 0.9190 & 0.9109 & 0.9182 & 0.9198 \\
        Class5  & 1.0000 & 0.9992 & 1.0000 & 0.9968 & 1.0000 & 1.0000 & 1.0000 & 1.0000 & 1.0000 & 1.0000 \\
        Class6  & \textbf{0.9968} & \textbf{0.9238} & \textbf{0.9175} & 0.9429 & 0.9556 & 0.9905 & 1.0000 & 0.9238 & 0.9968 & 1.0000 \\
        Class7  & \textbf{0.7544} & \textbf{0.6781} & 0.6987 & 0.6447 & 0.8855 & 0.8490 & 0.9269 & 0.8887 & 0.8967 & 0.8808 \\
        Class8  & \textbf{0.5162} & \textbf{0.4562} & 0.4765 & 0.4838 & 0.5462 & 0.5665 & 0.6345 & 0.5105 & 0.5721 & 0.5762 \\
        Class9  & \textbf{0.6723} & \textbf{0.7593} & 0.7295 & 0.8591 & 0.9356 & 0.8977 & 0.9163 & 0.9404 & 0.9122 & 0.9002 \\
        Class10 & \textbf{0.4462} & \textbf{0.5694} & \textbf{0.5062} & \textbf{0.4108} & 0.5522 & 0.4117 & 0.6204 & 0.4585 & 0.5440 & 0.5341 \\
        Class11 & \textbf{0.5902} & \textbf{0.8294} & 0.8408 & 0.8482 & \textbf{0.8147} & \textbf{0.5118} & \textbf{0.9053} & 0.7788 & 0.8180 & 0.8767 \\
        Class12 & 0.8806 & 0.8700 & 0.8937 & 0.8414 & 0.7719 & 0.8152 & 0.7874 & 0.8185 & 0.7073 & 0.7939 \\
        Class13 & 0.9172 & 0.8911 & 0.8301 & 0.8627 & 0.8497 & 0.8105 & 0.9020 & 0.9085 & 0.8824 & 0.9412 \\
        Class14 & 0.9904 & 1.0000 & 1.0000 & 1.0000 & 1.0000 & 1.0000 & 1.0000 & 0.9976 & 1.0000 & 1.0000 \\
        Class15 & 0.9877 & 0.9985 & 0.9969 & 0.9985 & 0.9985 & 1.0000 & 0.9815 & 0.9846 & 0.9969 & 1.0000 \\
       \hline
        OA      & 0.7830 & 0.8156 & 0.8113 & 0.8044 & 0.8452 & 0.8103 & \textbf{0.8739} & 0.8388 & 0.8431 & 0.8558 \\
        AA      & 0.8208 & 0.8433 & 0.8370 & 0.8339 & 0.8666 & 0.8385 & \textbf{0.8937} & 0.8617 & 0.8682 & 0.8811 \\
        Kappa   & 0.7658 & 0.8009 & 0.7962 & 0.7889 & 0.8327 & 0.7951 & \textbf{0.8637} & 0.8257 & 0.8305 & 0.8442 \\
        \hline
    \end{tabular}
    \label{summary_patch9_s10}
}
\end{table*}

\subsection{Impact of HSI Band Order on Cross-Architecture Validation}
To further evaluate our proposed approach generalization,  the existing two-branch CNN model is applied  by our reversed band order (DB2Li) data. As reported on Table~\ref{tab:architecture_performance}, we observed a 10\% improvement in OA under fewer samples (10) training conditions. Additionally, the dual-path fusion approach (DB3Li+DB4Li) achieved a competitive  OA and further validated that band order plays a critical role in classification performance across different network architectures. Moreover, our method's dual-path fusion (DB1Li+DB2Li) consistently outperforms individual orders, indicating that complementary spectral sequences provide additional discriminative power to improve classification performance.
\begin{table*}[ht!]
\centering
\caption{ {Architecture Performance Comparison: \\Two Branch-CNN model Applied our Proposed Different HSI Band order with LiDAR. Bold Data Represents the Best Method Experimental Results}}
\label{tab:architecture_performance}
\resizebox{\textwidth}{!}{ 
\begin{tabular}{|c|c|c|c|c|c|c|c|c|c|c|}
\hline
 & \multicolumn{2}{c|}{\textbf{Original Two Branch}} & \multicolumn{3}{c|}{\textbf{Two Branch (DB1Li, DB2Li) }} & \multicolumn{2}{c|}{\textbf{Two Branch (DB3, DB4, Li)}} & \multicolumn{3}{c|}{\textbf{Two Branch(DB3Li, DB4Li) }} \\
\hline
\textbf{Class} & \textbf{DB1+Li} & \textbf{DB2+Li} & \textbf{DB1Li} & \textbf{DB2Li} & \textbf{DB1Li+DB2Li} & \textbf{DB3+Li} & \textbf{DB4+Li} & \textbf{DB3Li} & \textbf{DB4Li} & \textbf{DB3i+DB4Li} \\
\hline

OA      & 0.7613 & 0.7314 & {0.8122} & 0.7920 &\textbf{ 0.8252} & 0.7526 & 0.7802 & {0.8390} & {0.8546} & \textbf{0.8664} \\
AA      & 0.8004 & 0.7776 & 0.8419 & 0.8261 & \textbf{0.8524} & 0.7898 & 0.8145 & {0.8603} & {0.8740} & \textbf{0.8863} \\
Kappa   & 0.7419 & 0.7105 & 0.7971 & 0.7752 & \textbf{0.8110} & 0.7326 & 0.7627 & 0.8259 & 0.8428 & \textbf{0.8555} \\
\hline
\end{tabular}
}
\end{table*}
Regarding the detailed class-wise performance comparison, we also have a detailed analysis. please refer to the supplementary document in the Github.

\subsection{Comparison with State-of-the-Art Fusion Methods}
Our method was compared with existing fusion approaches on Houston 2013 and Trento datasets. It achieved the highest accuracy in most classes and consistently outperformed prior models in OA and AA  
Table~\ref{tab:classification_results} compares various fusion methods applied to DB1Li+DB3Li based on both the Houston 2013 and Trento data sets and using standard training samples. The experimental results are better than other methods in terms of OA and AA. 
For Houston 2013 experimental result, our method achieves the best performance with an OA  0.9989 and AA of 0.9992,  consistently outperforming other methods. Trento Data experimental result has the OA 0.9972 and AA 0.9953 and outperforms the other method also.
Detailed experimental results for the class-wise result and analysis and  dataset can be found in the supplementary material.

   

\begin{table}[htbp]
    \centering
    \renewcommand{\arraystretch}{1.0}
     \caption{Classification Performance Comparison on Houston 2013 and Trento datasets.}
    \begin{tabular}{l|cc|cc} 
        \hline
        \multirow{2}{*}{Method} & \multicolumn{2}{c|}{Houston 2013} & \multicolumn{2}{c}{Trento} \\
        \cline{2-5}
        & OA & AA &  OA & AA \\
        \hline
        S2FL & 0.8681 & 0.8863  & 0.702 & 0.7647  \\
        CoSpace-L1 & 0.8752 & 0.8899  & 0.8348 & 0.8816  \\
        TwoBranchCNN & 0.9878 & 0.9136  & 0.8941 & 0.8910  \\
        MAHiDFNet & 0.8958 & 0.9653  & 0.9859 & 0.9755  \\
        FDCFNet & 0.9661 & 0.9692 & 0.9911 & 0.985  \\
        UCAFNet & 0.9680 & 0.9978 &  0.9926 & 0.9876  \\
        \textbf{Ours} & \textbf{0.9989} & \textbf{0.9992} & \textbf{0.9972} & \textbf{0.9953}  \\
        \hline
    \end{tabular}
   
    \label{tab:classification_results}
\end{table}

\subsection{Impact of Patch Size on Classification Performance}
\begin{figure}[!ht]
\centering
\includegraphics[width=7.5cm, height=4cm]{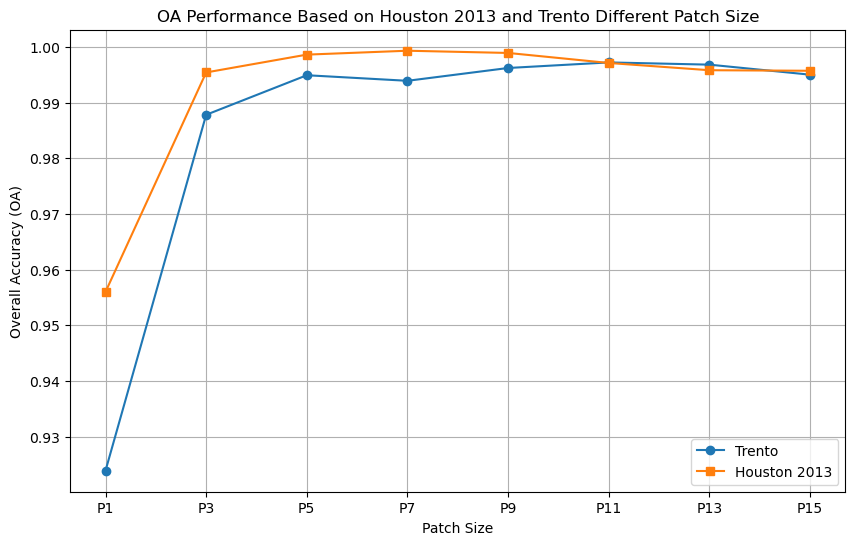}
\caption{OA and Patch Size Relationship based on Houston 2013 and Trento Datasets}
\label{fig4}
\end{figure}

Experiments on varying patch sizes (1 to 15) revealed dataset-dependent optimal configurations: Houston 2013  Best OA of 0.9993 at patch size 7; Trento  Best OA 0.9972 at patch size 11.
Fig.~\ref{fig4} illustrates larger patch sizes that improves spatial context utilisation but reduced sensitivity to band ordering.

\section{Conclusion}
In this study,  a novel architecture, HSLiNets, is introduced. The framework harnesses bidirectional reversed networks to fuse HSI and LiDAR features, leading to improvements in classification performance with different band orders. A comprehensive experimental results have demonstrated the proposed model improve the ability on capturing intricate multi-modal data characteristics and improving classification performance. Additionally, we have validated the model’s adaptability.
These findings underscore the effectiveness of our approach. Moving forward, we plan to further explore the fundamental principles behind the impact of spectral order on classification performance and computational complexity.

\textbf{Supplementary Material}: Additional experimental details and extended analysis are available on GitHub. 
\href{https://github.com/Judyxyang/HSLiNets/blob/main/Experiments/Experiemnt%20Summary.pdf}{HSLiNet\_Experimental\_Supplementary}.

\bibliographystyle{IEEEtran}

\bibliography{IEEEabrv, reference.bib}

\end{document}